\documentclass[a4paper, 10pt, conference]{ieeeconf}      

\usepackage{graphicx}
\usepackage{amssymb}
\usepackage{algorithmic}
\usepackage[ruled,linesnumbered]{algorithm2e}

\IEEEoverridecommandlockouts                              

\title{\LARGE \bf
Online Extrinsic Camera Calibration for Temporally Consistent IPM \\ Using Lane Boundary Observations with a Lane Width Prior}

\author{Jeong-Kyun Lee, Young-Ki Baik, Hankyu Cho, and Seungwoo Yoo
\thanks{All authors are with Qualcomm Korea YH, Seoul, South Korea
        {\tt\small \{ljeongky,ybaik,hankcho,yoos\}@qti.qualcomm.com}}%
}

\begin{document}

\maketitle
\thispagestyle{empty}
\pagestyle{empty}

\begin{abstract}

In this paper, we propose a method for online extrinsic camera calibration, \textit{i.e.}, estimating pitch, yaw, roll angles and camera height from road surface in sequential driving scene images. 
The proposed method estimates the extrinsic camera parameters in two steps: 1) pitch and yaw angles are estimated simultaneously using a vanishing point computed from a set of lane boundary observations, and then 2) roll angle and camera height are computed by minimizing difference between lane width observations and a lane width prior.
The extrinsic camera parameters are sequentially updated using extended Kalman filtering (EKF) and are finally used to generate a temporally consistent bird-eye-view (BEV) image by inverse perspective mapping (IPM).
We demonstrate the superiority of the proposed method in synthetic and real-world datasets.

\end{abstract}


\section{Introduction}

In recent years, researches for visual perception using a camera have been intensively conducted for the applications of ADAS (Advanced Driver Assistance System) and AD (Autonomous Driving).
Many of the researches focus on the detection of neighboring objects and driving environments, \textit{e.g.}, lane boundary detection~\cite{aly2008real,cao2019lane}, traffic sign detection~\cite{chira2010real,zhu2016traffic}, object detection and tracking~\cite{hu2019joint,song2015joint}, and so on, from input images captured by a front-facing camera.
Especially for road surface markings, inverse perspective mapping (IPM)~\cite{mallot1991inverse} is mainly utilized since they are more affected by camera perspective distortion. Given the geometric relationship between camera and road surface, \textit{i.e.}, extrinsic camera parameters, an input image can be converted into a bird-eye-view (BEV) image, thereby preserving the actual shapes of road surface markings and improving the performance of detecting them. Besides, the extrinsic camera parameters are widely exploited to estimate the distance of objects in monocular camera systems~\cite{song2014robust,stein2003vision} and improve the performance of object detection by generating enhanced features~\cite{liang2019multi}.

\begin{figure}[t]
    \small
	\begin{center}
	    \includegraphics[width=0.93\linewidth,height=83px]{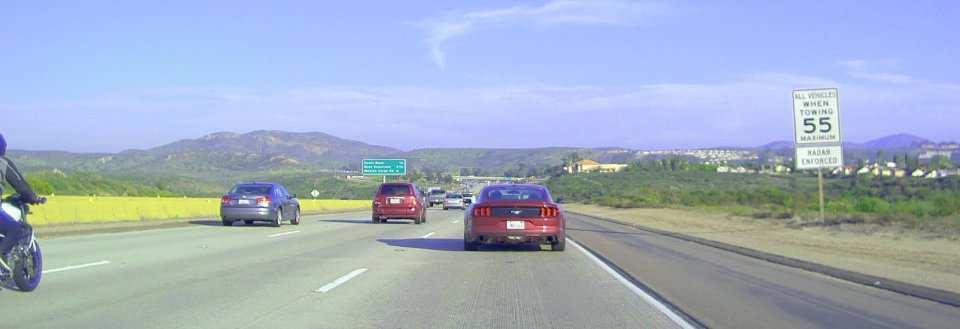} \\ 
	    (a) \\ 
	    \begin{tabular}{@{}c@{}c@{}}
		    \ \includegraphics[width=.46\linewidth,height=120px]{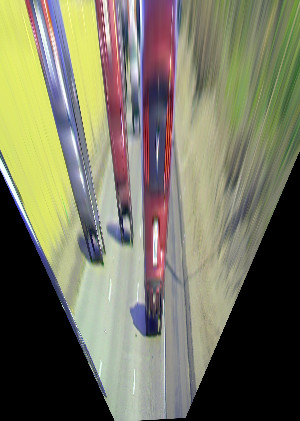} \ &
		    \includegraphics[width=.46\linewidth,height=120px]{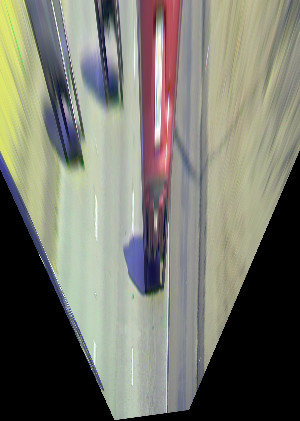} \\
		    (b) & (c)
		\end{tabular} 
	\end{center} 
	\caption{An example of improvement by our online extrinsic camera calibration for a driving scene image (a). (b) and (c) are BEV images by IPM using extrinsic camera parameters before and after applying our method, respectively.} 
	\label{fig:main} 
\end{figure}

The extrinsic camera parameters are usually calculated in advance of driving by using various patterns with rectangular/trapezoidal/parallelogram shape~\cite{chen2006vision,pang2013general} or manually annotated lines/vertices on lane markers~\cite{catala2006self,fung2003camera,wang2007research}.
However, the geometric relationship between camera and road surface gradually changes due to the non-persistence of factory default calibration and also it varies considerably as the camera is shaking while driving.
Accordingly, IPM cannot produce a good BEV image as shown in Fig.~\ref{fig:main}(b), so the extrinsic camera parameters should be repeatedly compensated while driving, but people cannot intervene in a calibration process while driving especially for safety reasons. In this context, extrinsic camera calibration should be automatically performed using observations from driving scene images.

There are several studies~\cite{jeong2016adaptive,zhang2014robust} that conduct online extrinsic camera calibration using image sequences of driving environments with camera motion variation.
They update the extrinsic camera parameters using camera motion estimated from visual odometry or a vanishing point (VP) of lane boundaries and result in BEV images showing parallel lane boundaries.
However, those works do not correct all the extrinsic camera parameters. They update only pitch and yaw angles, so, when roll angle and camera height change, they are at risk of producing BEV images where road surface fluctuates and scale, such as lane width and distance between objects, is inconsistent.

In this paper, we propose a method for online extrinsic camera calibration that estimates the geometric relationship such as pitch, yaw, roll angles and camera height from road surface in sequential driving scene images.
To the best of our knowledge, this is the first work of computing all the four extrinsic camera parameters concurrently in an online manner.
The proposed method estimates the extrinsic camera parameters in two stages that estimate 1) pitch and yaw angles and then 2) roll angle and camera height.
The pitch and yaw angles are estimated simultaneously using a VP computed from a set of lane boundary observations.
Then, given a lane width as a prior, roll angle and camera height are computed by minimizing difference between land width observations and the lane width prior.
The proposed method updates the extrinsic camera parameters using extended Kalman filtering (EKF)~\cite{kay1993fundamentals} in the sequential images so can produce temporally consistent IPM results as shown in Fig.~\ref{fig:main}(c).


\section{Overview}

\begin{figure}[t]
	\begin{center}
		\includegraphics[width=1\linewidth]{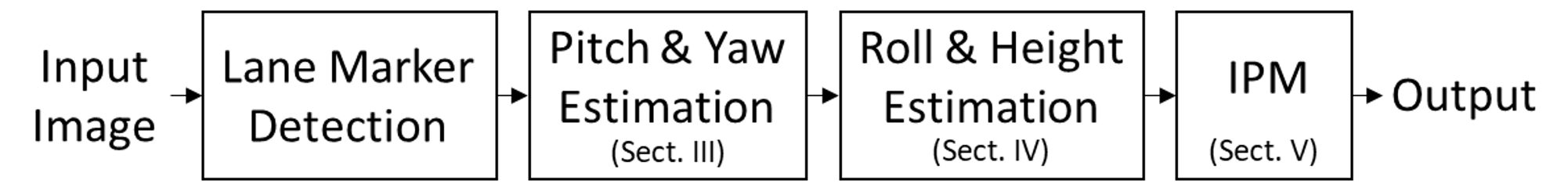}  
	\end{center} \vspace{-7pt}
	\caption{Overall procedure of the proposed method} 
	\label{fig:proc} 
\end{figure}

We propose a method for online extrinsic camera calibration, \textit{i.e.}, estimating pitch, yaw, roll angels and camera height from road surface, thereby producing temporally consistent IPM results.
We assume that camera is calibrated, road surface is flat, all the lane boundaries on road surface are parallel to each other, and lane widths are the same as a lane width prior.
Fig.~\ref{fig:proc} shows the overall procedure of the proposed method.
First, we extract lane boundary observations from an input image using a segmentation model based on fully convolutional networks~\cite{shelhamer2017fully}.
Since a VP of parallel lane boundaries depends only on pitch and yaw angles and is invariant to the change of roll angle and camera height, we find a VP from a set of parallel lane boundaries and estimate pitch and yaw angles using the VP (Sect.~\ref{sec:py}).
Then we compute roll angle and camera height minimizing the difference between lane width observations and the actual lane width given as a prior (Sect.~\ref{sec:rh}).
Finally, IPM is calculated using the updated extrinsic camera parameters (Sect.~\ref{sec:ipm}).

\section{Pitch and Yaw Angle Estimation} \label{sec:py}

\begin{figure}[t]
    \small
	\begin{center}
	    \begin{tabular}{@{}c@{}c@{}c@{}}
		    \includegraphics[width=.34\linewidth,height=85px]{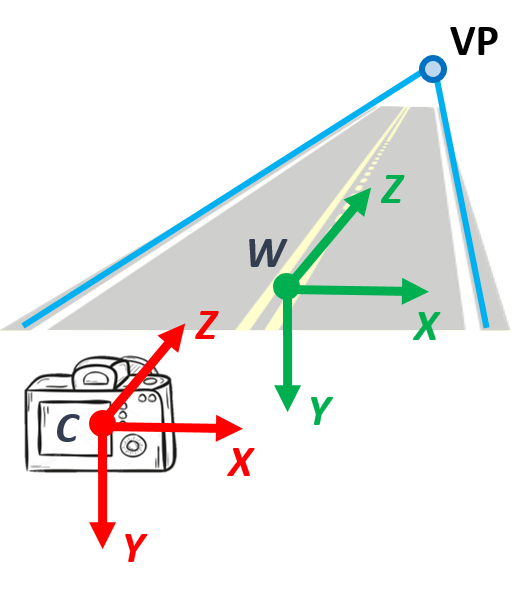} \ \ &
		    \includegraphics[width=.29\linewidth,height=85px]{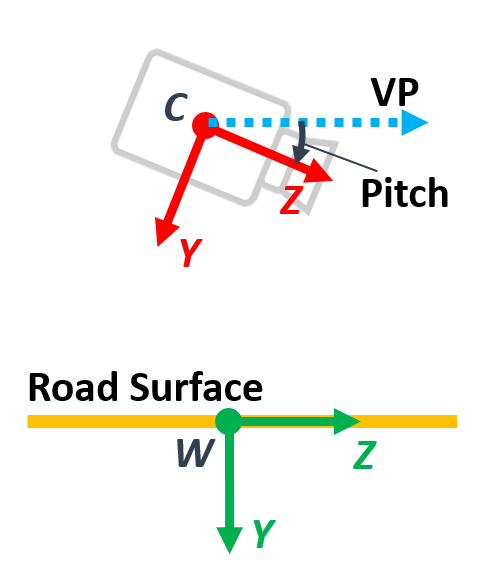} \ \ &
		    \includegraphics[width=.19\linewidth,height=85px]{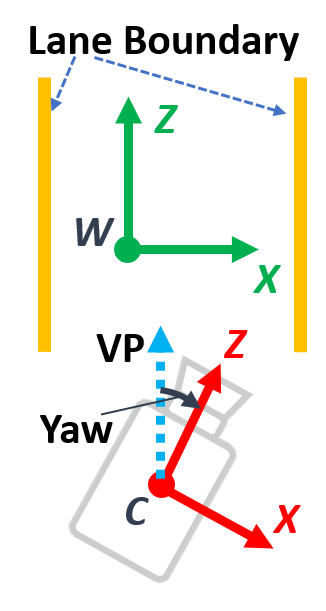}\\
		    (a) & (b) & (c)
		\end{tabular}
	\end{center} \vspace{-3pt}
	\caption{Pitch and yaw angle estimation. (a) Camera and world coordinate systems. (b) and (c) are the definitions of pitch and yaw angles, respectively.} 
	\label{fig:pitchyaw} 
\end{figure}

As in~\cite{catala2006self,wang2007research,zhang2014robust}, we translate the pitch and yaw angle estimation as finding a rotational relationship between camera and a VP of parallel lane boundaries on road surface as shown in Fig.~\ref{fig:pitchyaw}. $C$ and $W$ denote the camera and world coordinate systems, respectively. Let us define the $z$-axis of $W$ as the direction of VP, \textit{i.e.}, VD (Vanishing Direction). Then the pitch and yaw angles can be defined as angles between the forward direction of camera and VD as shown in Fig.~\ref{fig:pitchyaw}(b) and Fig.~\ref{fig:pitchyaw}(c).
We employ the robust VP estimation method of \cite{lee2019joint} based on the Gaussian sphere theory and RANSAC~\cite{fischler1981random} since lane boundary observations can be noisy. After the pitch and yaw angles are initialized using the VP, they are estimated in the sequential images by EKF.

\subsection{Gaussian Sphere} \label{sec:knowledge}

\begin{figure}[t]
	\begin{center}
		\includegraphics[width=0.94\linewidth]{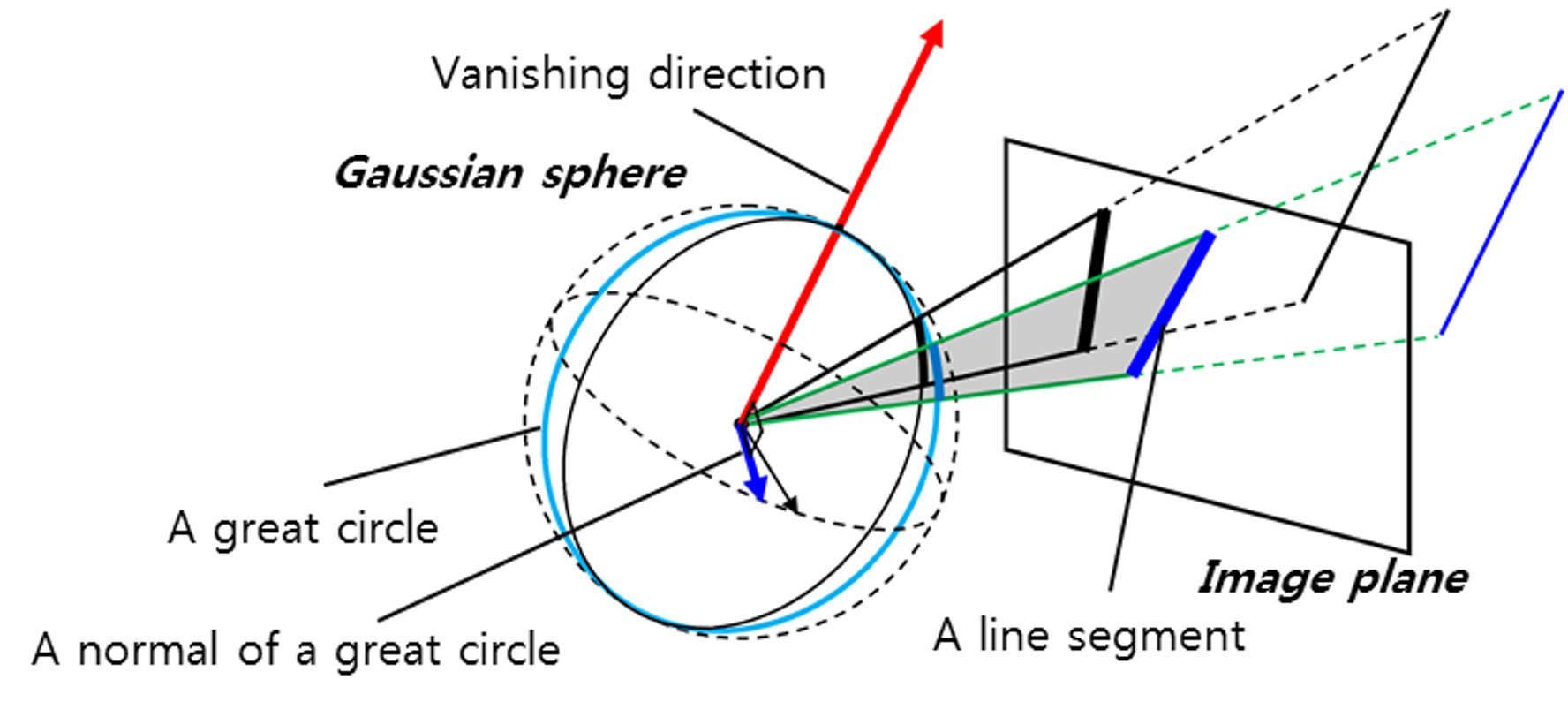}  
	\end{center} \vspace{-3pt}
	\caption{Description of the Gaussian sphere} 
	\label{fig:gausssphere} 
\end{figure}

In the pinhole camera model, a unit sphere centered at the principal point of the camera is called the Gaussian sphere. As shown in Fig.~\ref{fig:gausssphere}, a great circle is the intersection of the Gaussian sphere and the plane determined by a line on the image and the principal point. As the parallel lines meet at VP when projected on the image plane, the great circles corresponding to parallel lines have an intersection point on the Gaussian sphere, and the direction from the principal point to the intersection point becomes VD. VD is the normal vector of the plane determined by all the normals of the great circles (NGCs), which we refer to as the NGC-VD orthogonality. The orthogonality is the same as the line-VP incidence in the image plane, \textit{i.e.}, parallel lines in the image plane are incident to VP.

\subsection{Vanishing Point Estimation} \label{sec:vpest}

\begin{algorithm} [t] 
	\small
	\caption{RANSAC-based VP estimation}
	\label{alg:ransac}
	\begin{algorithmic}[1]
		\REQUIRE $\mathcal{L}$
		\ENSURE $\mathbf{v}$ and $\mathcal{L}_c$
		\STATE $\mathcal{L}_c = \{ \varnothing \}, s_{max} = 0$
		\FOR{$i = 1$ \TO $N_{loop}$}
			\STATE Compute a VP hypothesis $\mathbf{v}_{i}$ from two randomly samp- led lines $\{\mathbf{l}_j, \mathbf{l}_k\} \in \mathcal{L}$ using (\ref{eq:vphypcomput})
			\STATE Compute a score value $s_{i} = S(\mathbf{v}_i, \mathcal{L})$ using (\ref{eq:score})
			\IF{$s_i > s_{max}$}
				\STATE $s_{max} = s_i$, $\mathcal{L}_c = \{ \mathbf{l} | \mathbf{l} \in \mathcal{L}, \theta(\mathbf{v}_i,\mathbf{l}) < \theta_{th} \}$
			\ENDIF
		\ENDFOR
		\STATE Compute a VP $\mathbf{v}$ from $\mathcal{L}_c$ using (\ref{eq:vpinit})
	\end{algorithmic}
\end{algorithm}

We assume that a set of lines representing lane boundaries is given. The set usually includes some noisy lines or outliers so we filter out the outliers using RANSAC~\cite{fischler1981random} and then estimate a VP robust to noisy lines. When a set of line segments $\mathcal{L}$ is given, the RANSAC process can be described as Algorithm~\ref{alg:ransac}. 

In Algorithm~\ref{alg:ransac}, a VP hypothesis $\mathbf{v}_i$ is computed from two randomly sampled line segments $\{\mathbf{l}_j, \mathbf{l}_k\} \subset \mathcal{L}$ as follows:

\begin{equation} \label{eq:vphypcomput}
	\mathbf{v}_i = \mathbf{l}_j \times \mathbf{l}_k.
\end{equation}

\begin{figure}[t]
	\begin{center}
		\includegraphics[width=0.8\linewidth]{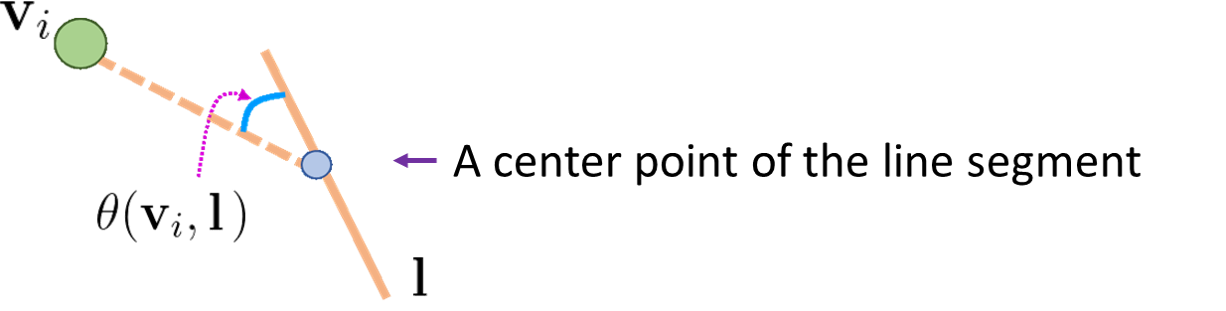}  
	\end{center} \vspace{-3pt}
	\caption{Description of Rother's score function~\cite{rother2002new}} 
	\label{fig:rother} 
\end{figure}

Then a score value $s_i$ of the VP hypothesis $\mathbf{v}_i$ is computed using the score function of Rother~\cite{rother2002new}. Fig.~\ref{fig:rother} shows two constraints to be considered for computing $s_i$: for each line segment $\mathbf{l} \in \mathcal{L}$, 1) an angle $\theta(\mathbf{v}_i,\mathbf{l})$ between $\mathbf{l}$ and an imaginary line including $\mathbf{v}_i$ and the center point of $\mathbf{l}$, and 2) the length of $\mathbf{l}$, $l_\mathbf{l}$. Now the score function is defined as

\begin{equation} \label{eq:score}
	S(\mathbf{v}_i, \mathcal{L}) = \sum_{\mathbf{l} \in \mathcal{L}} \left[ \lambda_1 \left( 1 - \frac{\theta(\mathbf{v}_i,\mathbf{l})}{\theta_{th}} \right) + \lambda_2 \frac{l_\mathbf{l}}{l_{m}} \right],
\end{equation}

\noindent where $\lambda_1$ and $\lambda_2$ are weighting values and set to 0.8 and 0.2, respectively. $\theta_{th}$ is a threshold value and set to $0.7^\circ$. $l_{m}$ is the length of the longest line segment in $\mathcal{L}$. When $\theta(\mathbf{v}_i,\mathbf{l})$ is not less than $\theta_{th}$, $\mathbf{l}$ is not included in the score computation.

Consequently, a set of line segments $\mathcal{L}_c$ with the highest score is clustered by the RANSAC process.
The clustered lines are used to compute an optimal VP.
To obtain the optimal VP, we utilize the NGC-VD orthogonality as mentioned in Sect.~\ref{sec:knowledge}. An NGC $\mathbf{n}$ of each line $\mathbf{l}$ is computed by

\begin{equation} \label{eq:ngc}
	\mathbf{n} = (\mathbf{K}^{-1}\mathbf{p}_{1}) \times (\mathbf{K}^{-1}\mathbf{p}_{2}),
\end{equation}

\noindent where $\mathbf{K}$ is an intrinsic camera matrix, and $\mathbf{p}_1$ and $\mathbf{p}_2$ are the end points of $\mathbf{l}$. Then the orthogonality equation of $\mathbf{v}$ and the NGCs is as follows:

\begin{equation} \label{eq:vpinit}
	\mathbf{A} \mathbf{v} = 0, \ \mathrm{where} \ \mathbf{A} = [\cdots, \mathbf{n}, \cdots]^{\top}.
\end{equation}

The over-determined system of the linear equations can be solved easily by singular value decomposition (SVD). Actually, $\mathbf{v}$ computed in (\ref{eq:vpinit}) is a VD vector that is projected into a VP on the image plane by $\mathbf{K}$, \textit{i.e.}, $\mathbf{v}_d = \mathbf{K}^{-1} \mathbf{v}_p$ where $\mathbf{v}_d$ and $\mathbf{v}_p$ are VD and VP, respectively, but those are identical, so VD would be written as VP from now on.

\subsection{Pitch and Yaw Angle Initialization} \label{sec:py_sub}

The pitch and yaw angles are denoted by $\theta$ and $\phi$, respectively. 
A rotation matrix calculated by the pitch and yaw angles, \textit{i.e.}, a transformation matrix from world coordinates to camera coordinates, is represented by $\mathbf{R}_{CW}(\theta, \phi)$ and a direction vector of $z$-axis in the world coordinate system $W$ is denoted by $\mathbf{d}_{W_Z} = [0, 0, 1]^{\top}$.
Then $\mathbf{d}_{W_Z}$ and $\mathbf{v}$ have the following relationship,

\begin{equation}
	\mathbf{v} = \mathbf{R}_{CW}(\theta, \phi) \mathbf{d}_{W_Z}.
\end{equation}

We can decompose the rotation matrix into two rotation matrices of $\theta$ and $\phi$ as follows.
\begin{eqnarray}
\mathbf{R}_{CW}(\theta, \phi) & = & \mathbf{R}(\theta) \mathbf{R}(\phi) \nonumber \\
& = & \left[ \begin{array}{ccc} 1 & 0 & 0 \\ 0 & c_\theta & -s_\theta \\ 0 & s_\theta & c_\theta \end{array}\right]
\left[ \begin{array}{ccc} c_\phi & 0 & s_\phi \\ 0 & 1 & 0  \\ -s_\phi & 0 & c_\phi \end{array}\right]
\end{eqnarray}

\noindent where $c_\theta$ and $s_\theta$ ($c_\phi$ and $s_\phi$) are the values of cosine and sine functions of $\theta$ ($\phi$).
Then $\theta$ and $\phi$ are initialized from $\mathbf{v}$ as follows.
\begin{eqnarray}
\theta & = &  \mathrm{atan2} \left( -v_y, v_z\right) \ \mathrm{and} \label{eq:pitchinit} \\
\phi & = & \mathrm{atan2} \left( v_x, v_z\right), \label{eq:yawinit}
\end{eqnarray}

\noindent where $\mathbf{v} = [v_x, v_y, v_z]^\top$ and atan2$(y,x)$ is the 2-argument arctangent function. 

\subsection{Pitch and Yaw Angle Estimation by EKF} \label{sec:pyekf}

We employ EKF~\cite{kay1993fundamentals} for the pitch and yaw angle estimation in an image sequence.
A constant angular velocity model is adopted to model pitch and yaw angle variation while driving.
Accordingly, a state vector $\mathbf{x}_{PY}$ and a system model $\mathbf{f}_{PY}$ for the pitch and yaw angle estimation are defined as
\begin{eqnarray}
\mathbf{x}_{PY} & = & \left[ \theta, \phi, \omega_\theta, \omega_\phi  \right] ^\top \ \mathrm{and} \label{eq:state_py} \\
\mathbf{f}_{PY}(\mathbf{x}_{PY}) & = & \left[ \begin{array}{c} \theta + \omega_\theta \Delta t \\ \phi + \omega_\phi \Delta t \\ \omega_\theta \\ \omega_\phi \end{array} \right] + \mathbf{w}_{PY}, \label{eq:sys_py}
\end{eqnarray}

\noindent where $\omega_\theta$ and $\omega_\phi$ are the angular velocity of pitch and yaw angles and $\mathbf{w}_{PY} = [0,0,w_\theta,w_\phi]^\top$ is a noise variable of the system model with the normal distribution $\mathcal{N}(\mathbf{0}, \mathbf{W}_{PY})$.
Using the NGC-VD orthogonality, a measurement model $h_{PY}$ for the pitch and yaw angle estimation is defined as

\begin{equation} \label{eq:meas_py}
	h_{PY}^\mathbf{n}(\mathbf{x}_{PY}) = \mathbf{d}_{W_Z}^\top \mathbf{R}_{CW}(\theta,\phi)^\top  \mathbf{n} + q_{PY},
\end{equation}

\noindent where $\mathbf{n}$ is an NGC of $\mathbf{l} \in \mathcal{L}_c$ and $q_{PY}$ is a noise variable of the measurement model with the normal distribution~$\mathcal{N}(0,Q_{PY})$.

The state vector $\mathbf{x}_{PY}$ are estimated by EKF as follows:
For simplicity, $\mathbf{x}_{PY}$ is abbreviated as $\mathbf{x}$.
Then let $\mathbf{x}_{t-1}$ and $\mathbf{P}_{t-1}$ denote the state estimate and its covariance at time $t-1$.
The prediction of the state $\hat{\mathbf{x}}_t$ and its covariance $\hat{\mathbf{P}}_t$ is calculated as follows.
\begin{eqnarray} 
\hat{\mathbf{x}}_t & = & \mathbf{f}_{PY}(\mathbf{x}_{t-1}) \ \mathrm{and} \label{eq:ekf_state} \\
\hat{\mathbf{P}}_t & = & \mathbf{F}_t\mathbf{P}_{t-1}\mathbf{F}_t^\top + \mathbf{W}_t,\label{eq:ekf_cov}    
\end{eqnarray}

\noindent where $\mathbf{F}_t$ is a Jacobian of the system model.
For all the inliers in $\mathcal{L}_c$, the residuals of the measurements are computed using (\ref{eq:meas_py}) and concatenated as follows.

\begin{equation} \label{eq:ekf_meas_py}
    \mathbf{y}_t = -[\cdots, \ h_{PY}^\mathbf{n}(\hat{\mathbf{x}}_t), \ \cdots]^\top
\end{equation}

\noindent The Kalman gain $\mathbf{G}_t$ for update is computed as follows.

\begin{equation}
\mathbf{G}_t = \hat{\mathbf{P}}_{t} \mathbf{H}^\top_t \mathbf{S}_t^{-1}, \ \mathrm{where} \ \mathbf{S}_t = \mathbf{H}_t \hat{\mathbf{P}}_t \mathbf{H}^\top_t + \mathbf{Q}_t.
\end{equation}

\noindent Here, $\mathbf{H}_t$ and $\mathbf{Q}_t$ denote a Jacobian of the measurement model, \textit{i.e.}, $\mathbf{H}_t = \left[ \cdots, \ \frac{\partial h_{PY}^\mathbf{n}}{\partial \mathbf{x}} \vert _{\hat{\mathbf{x}}_t}^\top, \ \cdots \right]^\top$, and a measurement noise covariance at time $t$, respectively.
From the residuals and the Kalman gain, the predicted state and covariance are updated as follows.
\begin{eqnarray}
\mathbf{x}_t & = & \hat{\mathbf{x}}_t + \mathbf{G}_t \mathbf{y}_{t} \ \mathrm{and} \label{eq:ekf_updx} \\
\mathbf{P}_t & = & (\mathbf{I} -  \mathbf{G}_t \mathbf{H}_t ) \hat{\mathbf{P}}_t \label{eq:ekf_updp},
\end{eqnarray}

\noindent where $\mathbf{I}$ is an identity matrix.


\section{Roll Angle and Camera Height Estimation} \label{sec:rh}

\begin{figure}[t]
    \small
	\begin{center}
	    \begin{tabular}{@{}c@{}c@{}c@{}}
		    \includegraphics[width=.35\linewidth,height=54px]{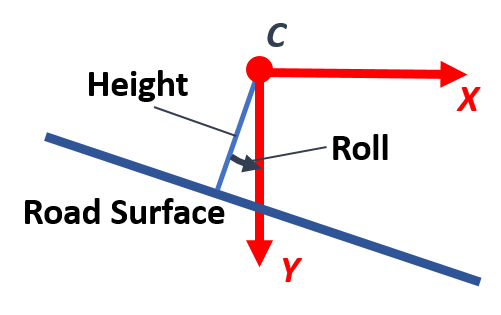} \ \ &
		    \includegraphics[width=.315\linewidth,height=75px]{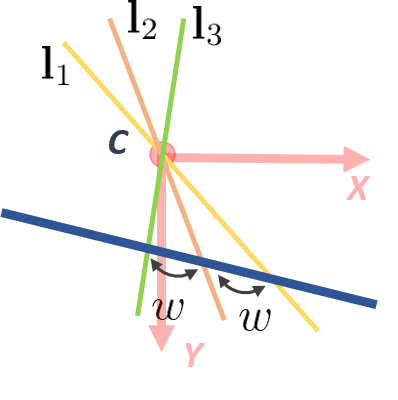} \ \ &
		    \includegraphics[width=.29\linewidth,height=69px]{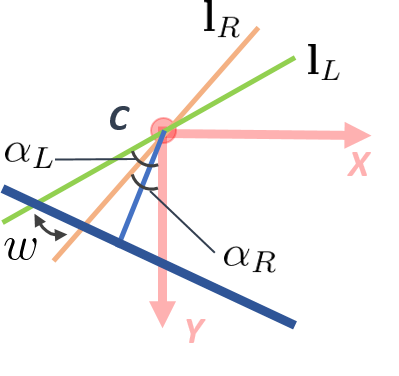}\\
		    (a) & (b) & (c)
		\end{tabular}
	\end{center} \vspace{-3pt}
	\caption{Roll angle and camera height estimation. (a) Definition of roll angle and camera height. (b) Definition of lane width. (c) Definition of line angles.}  \vspace{-10pt}
	\label{fig:rollheight} 
\end{figure}

In comparison with~\cite{fung2003camera} using the lane width prior and manually annotated 3D vertices on road surface as observations, it is more complicated to calibrate roll angle and camera height using the 2D projection of lane boundaries as observations due to the lack of geometric information occurred from projective properties and the nonlinear geometric relationship between the observed lane boundaries and the extrinsic camera parameters. For simplification, we assume that the pitch and yaw angles are already corrected by in Sect.~\ref{sec:py}.
Then the roll angle and camera height estimation can be thought as computing their approximations on $xy$-plane as shown in Fig.~\ref{fig:rollheight}. By projecting road surface and lines $\mathbf{l}$ onto $xy$-plane, we can estimate the roll angle and camera height values at which the distances between intersection points of road surface and lines should be equal to the lane width prior $w_p$.

\subsection{Roll Angle and Camera Height Initialization} \label{eq:rhinit}

We first find line pairs $\{\mathbf{l}_L, \mathbf{l}_R\} \subset \mathcal{L}_c$ on the both sides of a lane and compute the line angles ($\alpha_L$ and $\alpha_R$) with respect to $y$-axis as shown in Fig.~\ref{fig:rollheight}(c).
We let $\psi$ and $h$ denote roll angle and camera height, respectively.
Then a lane width can be calculated as 
\begin{equation} \label{eq:west}
    w_{\mathbf{l}_L,\mathbf{l}_R} (\psi, h) = h(\tan{(\alpha_L - \psi)} - \tan{(\alpha_R - \psi)}).
\end{equation}

\noindent An energy function for the roll angle and camera height estimation is defined as
\begin{equation} \label{eq:whenergy}
    E_{RH} (\psi, h) =  \sum_{(\mathbf{l}_L, \mathbf{l}_R) \in \mathcal{L}_c}  C_{\mathbf{l}_L,\mathbf{l}_R} (\psi, h),
\end{equation}
where
\begin{equation} \label{eq:whcost}
    C_{\mathbf{l}_L,\mathbf{l}_R} (\psi, h) =  w_p - w_{\mathbf{l}_L,\mathbf{l}_R} (\psi, h).
\end{equation}

$\psi$ and $h$ are initialized by an exhaustive search within a limited range and then optimized by minimizing the energy function of (\ref{eq:whenergy}) using the Gauss-Newton method.
It is noteworthy that at least two lanes should be considered since the roll angle is estimated from the equality of a pair of lane width observations.

\subsection{Roll Angle and Camera Height Estimation by EKF} \label{sec:rhekf}

In a similar manner to Sect.~\ref{sec:pyekf}, we use constant angular and linear velocity models for temporally consistent estimation of roll angle and camera height.
Accordingly, a state vector $\mathbf{x}_{RH}$ and a system model $\mathbf{f}_{RH}$ for the roll angle and camera height estimation are defined as
\begin{eqnarray}
\mathbf{x}_{RH} & = & \left[ \psi, h, \omega_\psi, v_h  \right] ^\top \ \mathrm{and} \\
\mathbf{f}_{RH}(\mathbf{x}_{RH}) & = & \left[ \begin{array}{c} \psi + \omega_\psi \Delta t \\ h + v_h \Delta t \\ \omega_\psi \\ v_h \end{array} \right] + \mathbf{w}_{RH}.
\end{eqnarray}
Here, $\omega_\psi$ and $v_h$ are the angular velocity of roll angle and the linear velocity of camera height, and $\mathbf{w}_{RH} = [0,0,w_\psi,v_h]^\top$ is a noise variable of the system model with the normal distribution $\mathcal{N}(\mathbf{0}, \mathbf{W}_{RH})$.
A measurement model $\mathbf{h}_{RH}(\psi,h)$ for the roll angle and camera height estimation is the same as (\ref{eq:whcost}).
Finally, roll angle and camera height are estimated by EKF using Eqs.~(\ref{eq:ekf_state})-(\ref{eq:ekf_updp}).


\section{Inverse Perspective Mapping} \label{sec:ipm}

Finally, temporally consistent IPM is possible with the extrinsic camera parameter estimates $\theta$, $\phi$, $\psi$, and $h$.
A homography matrix $\mathbf{H}_{WC}$ from the camera coordinates to the world (or ground) coordinates is calculated as follows.
\begin{equation} \label{eq:homography}
\mathbf{H}_{WC} = \left[ \begin{array}{ccc} a_X & 0 & \frac{b_X a_X}{2} \\ 0 & -a_Z & b_Z a_Z \\ 0 & 0 & 1 \end{array}\right] \left[\begin{array}{c} \mathbf{R}_{r_1} \\ \mathbf{R}_{r_3} \\ \frac{1}{h}\mathbf{R}_{r_2} \end{array}\right] \mathbf{K}^{-1},
\end{equation}
where $a_X$ and $a_Z$ are the scale parameters that determine the resolution of a BEV image, $b_X$ and $b_Z$ are the size parameters of $X$-axis and $Z$-axis in the world coordinate system, and $\mathbf{R}_{r_i}$ is the $i$-th row of the matrix $\mathbf{R}(\theta,\phi,\psi)$ which is computed by
\begin{equation}
\mathbf{R}(\theta,\phi,\psi)  =  \left[ \begin{array}{ccc} \cos \psi & \sin \psi & 0 \\ - \sin \psi & \cos \psi & 0 \\ 0 & 0 & 1 \end{array}\right] \mathbf{R}_{CW}(\theta, \phi)^\top.
\end{equation}
Using (\ref{eq:homography}), an input image can be converted into a BEV image as shown in Fig.~\ref{fig:main}(c).


\section{Experimental Results}

\subsection{Experimental Setup}

\paragraph{Lane Boundary Detection}

\begin{figure}[t]
    \small
	\begin{center}
	    \begin{tabular}{@{}c@{}c@{}}
		    \includegraphics[width=.49\linewidth,height=60px]{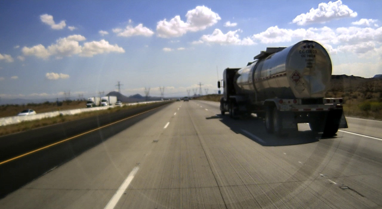} \ &
		    \includegraphics[width=.49\linewidth,height=60px]{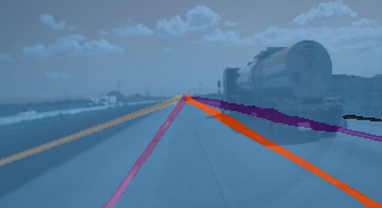}
		    \\
		    (a) & (b)
		\end{tabular} 
	\end{center} 
	\caption{Result of the segmentation model. (a) An input image. (b) Segmentation model output where each color represents each semantic lane boundary instance.}
	\label{fig:lanemarker_vis} 
\end{figure}

To obtain lane boundary observations, we employ a segmentation model based on fully convolutional networks~\cite{shelhamer2017fully}. The segmentation model performs multi-class segmentation where each semantic lane boundary has its own class. We use ResNet-18~\cite{He_2016} as a backbone and the standard SGD method~\cite{bottou_10} with a cross entropy loss for training.
As shown in Fig.~\ref{fig:lanemarker_vis}, the segmentation model provides each semantic lane boundary instance by generating a pixel-wise probability map of each semantic lane. Subsequently, for each lane boundary whose probability is larger than 0.5, we extract vertices by searching through the mean position of on-pixels and then sample representative vertices for the next use.

\paragraph{Dataset}
The proposed method is evaluated using several synthetic and real-world datasets as shown in Fig.~\ref{fig:syn_res} and Fig.~\ref{fig:real_res}.
For the synthetic dataset, we synthesized an image sequence of 300 frames at a resolution of $1920 \times 1020$ pixels where there are 5 lanes, \textit{i.e.}, 6 lane boundaries, on a planar road surface and was captured with the change of extrinsic camera parameters.
We also captured two image sequences (\textit{Test1} and \textit{Test2}) with a resolution of $1920 \times 1020$ pixels in the real-world scenes.
In all the datasets, the lane width prior $w_p$ is set to 3.7 meters which are the standard lane width in the U.S. highway system.

\subsection{Evaluation in the Synthetic Dataset}

\begin{figure}[t]
    \small
	\begin{center}
	    \begin{tabular}{@{}c@{}c@{}}
		    \includegraphics[width=.62\linewidth,height=68px]{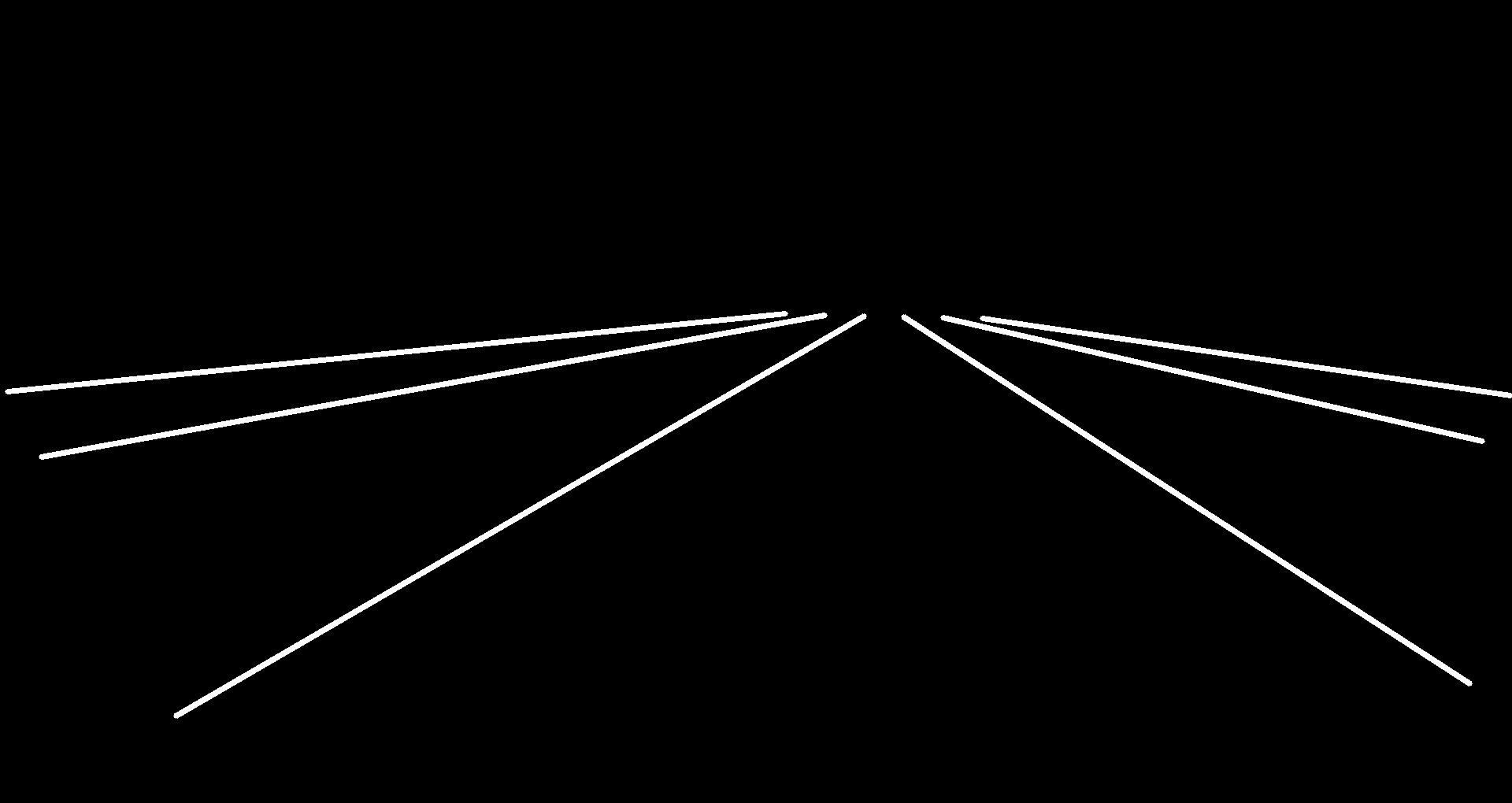} \ &
		    \includegraphics[width=.35\linewidth,height=68px]{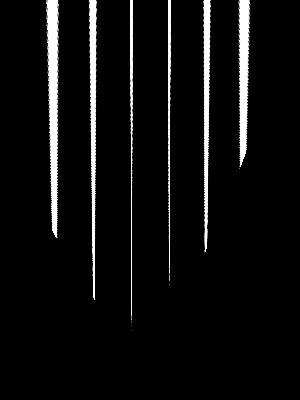} \\
		\end{tabular} \\
		Frame \#15 \\ \vspace{5pt}
		\begin{tabular}{@{}c@{}c@{}}
		    \includegraphics[width=.62\linewidth,height=68px]{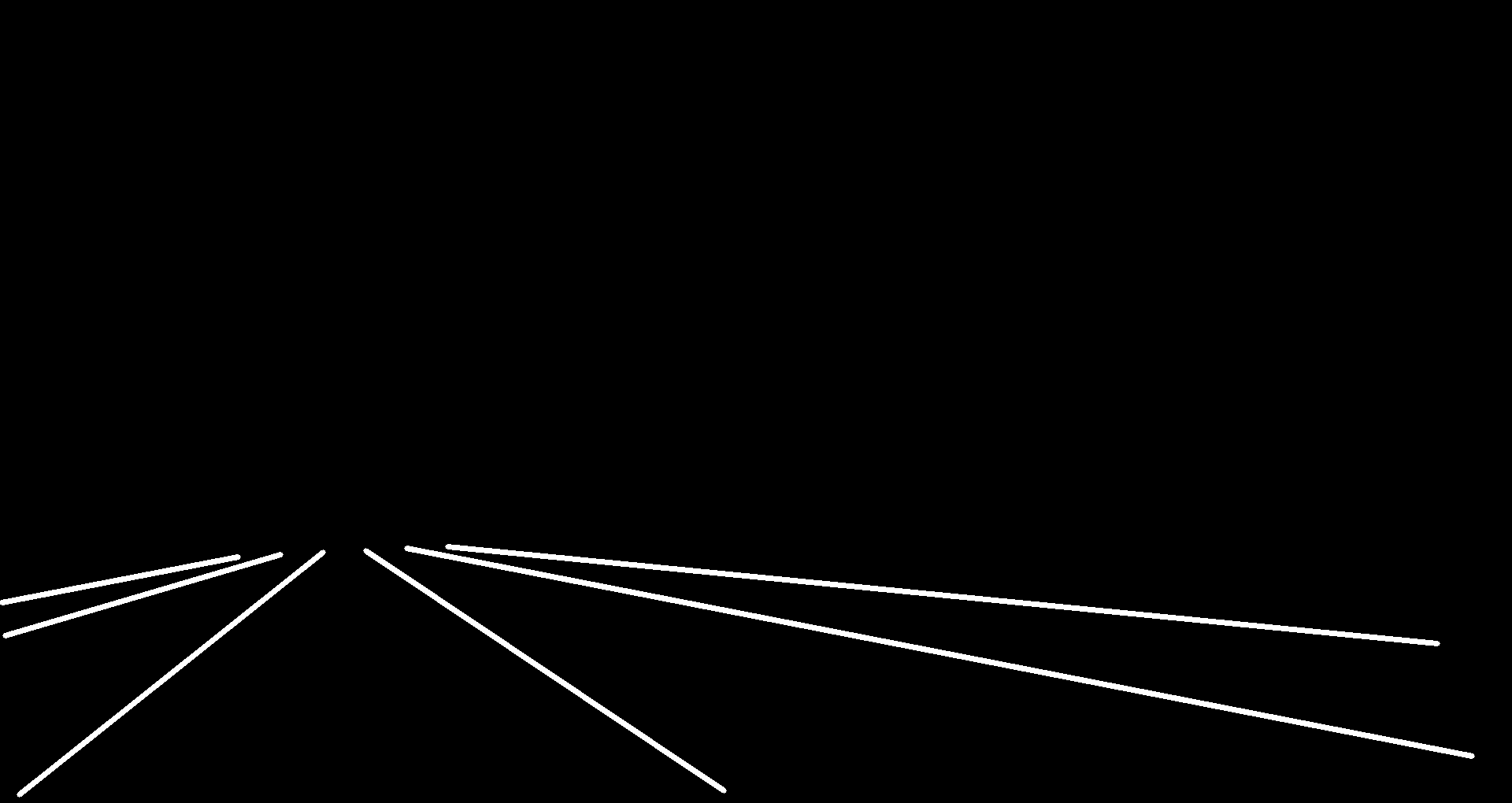} \ &
		    \includegraphics[width=.35\linewidth,height=68px]{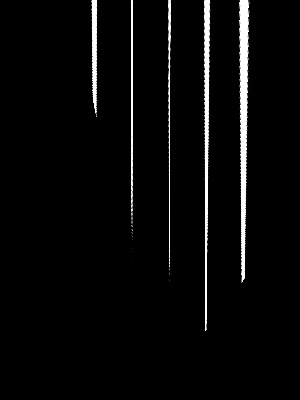} \\
		\end{tabular} \\ 
		Frame \#214 \\ \vspace{-5pt}
	\end{center} 
	\caption{Result of our online extrinsic camera calibration without lane boundary detection in the synthetic scenes with the noise of $\sigma^2 = 1$. Left and right images show the input images and their BEV images produced by the proposed method, respectively.} 
	\label{fig:syn_res} 
\end{figure}

\begin{table}[t]
    \caption{RMSE of the extrinsic camera parameter estimates on the synthetic dataset}
    \begin{center}
        \begin{tabular}{ccccc}
            \hline \noalign{\smallskip}
            Noise & Pitch (deg.) & Yaw (deg.) & Roll (deg.) & Height (cm)  \\ \hline \noalign{\smallskip}
            $\sigma^2 = 0.5$ & 0.037 & 0.104 & 0.059 & 0.60 \\ 
            $\sigma^2 = 1.0$ & 0.039 & 0.105 & 0.067 & 0.69 \\
            $\sigma^2 = 2.0$ & 0.045 & 0.111 & 0.077 & 0.83 \\
            $\sigma^2 = 4.0$ & 0.056 & 0.120 & 0.090 & 1.03 \\
            $\sigma^2 = 9.0$ & 0.060 & 0.141 & 0.114 & 1.40 \\\hline
        \end{tabular}
    \end{center}
    \label{tab:syn_res}
\end{table}

Due to the lack of ground truth for quantitative evaluation in a real-world scenario, we generated the synthetic dataset as shown in the left images of Fig.~\ref{fig:syn_res} where lane boundaries are represented by white lines.
In detail, we generated equally spaced points (at 30 pixel intervals) on each lane boundary and obtained line segments by randomly sampling and pairing two points. We acquired at most 30 pairs of the line segments per lane. The synthetic dataset included average 408 line segments per frame. Then Gaussian noises with variance of $\sigma^2 = 0.5$, 1, 2, 4, and 9 in pixels are added at the end points of the line segments.
We conducted the Monte Carlo experiments where the proposed method ran 100 times for each noise variance and was evaluated by measuring the root mean square errors (RMSE) of the pitch, yaw, roll, and height estimates.

Table~\ref{tab:syn_res} shows the results evaluated in the synthetic dataset. The RMSEs increase in proportion to the noise variance but the RMSEs of the rotational angles and the camera height estimates are less than 0.2 degrees and 2 centimeters, respectively, even in the presence of the severe noise of $\sigma^2 = 9$ in pixels. In addition, the BEV images produced by the proposed method for two frames are temporally consistent despite the motion variation and noise added. In conclusion, the proposed method performed quite well on the synthetic dataset.

\subsection{Evaluation in the Real-world Dataset}

\begin{figure}[t]
    \small
	\begin{center}
	    \begin{tabular}{@{}c@{}c@{}c@{}c@{}}
	        \multicolumn{2}{@{}c@{}}{\includegraphics[width=0.48\linewidth,height=42px]{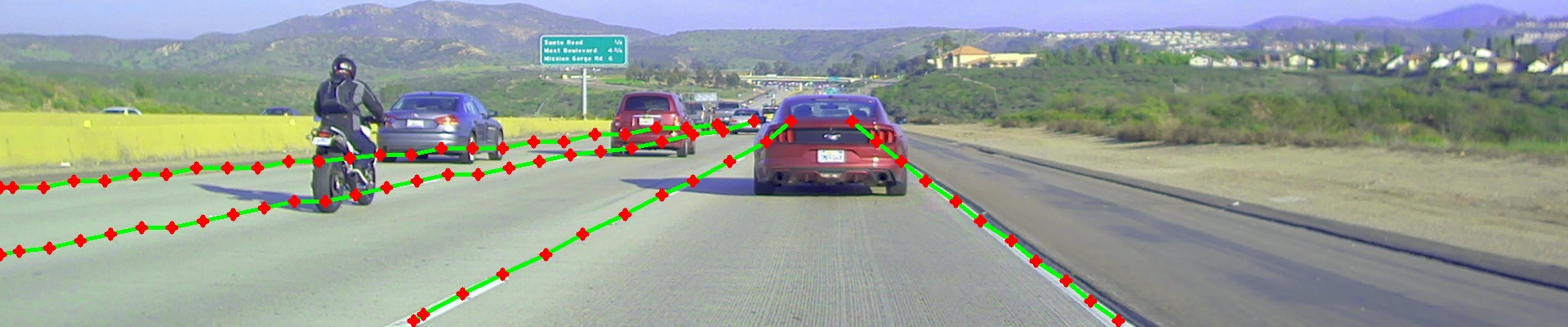}} \ &
	        \multicolumn{2}{@{}c@{}}{\includegraphics[width=0.48\linewidth,height=42px]{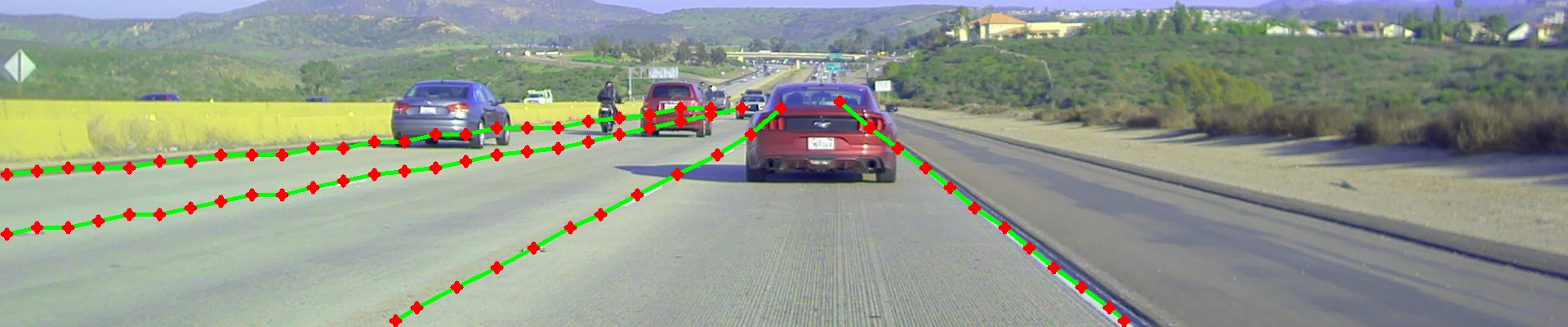}} \\
		    \includegraphics[width=.235\linewidth,height=72px]{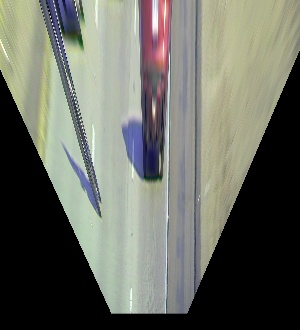} \ &
		    \includegraphics[width=.235\linewidth,height=72px]{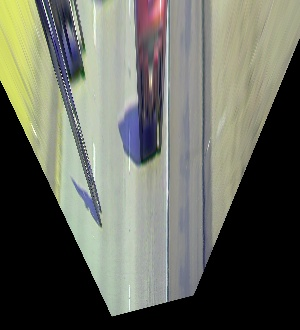} \ &
		    \includegraphics[width=.235\linewidth,height=72px]{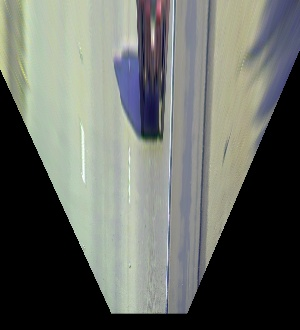} \ &
		    \includegraphics[width=.235\linewidth,height=72px]{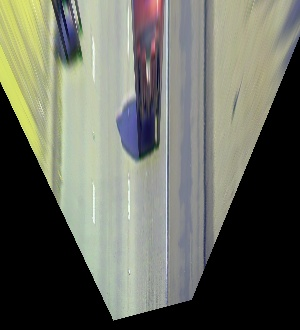}\\
		    \multicolumn{2}{@{}c@{}}{Frame \#55} &
		    \multicolumn{2}{@{}c@{}}{Frame \#265} \\
		    \multicolumn{4}{@{}c@{}}{\textit{Test1} dataset} \\
		    \rule{0pt}{0.4ex} & & & \\
		    \multicolumn{2}{@{}c@{}}{\includegraphics[width=0.48\linewidth,height=42px]{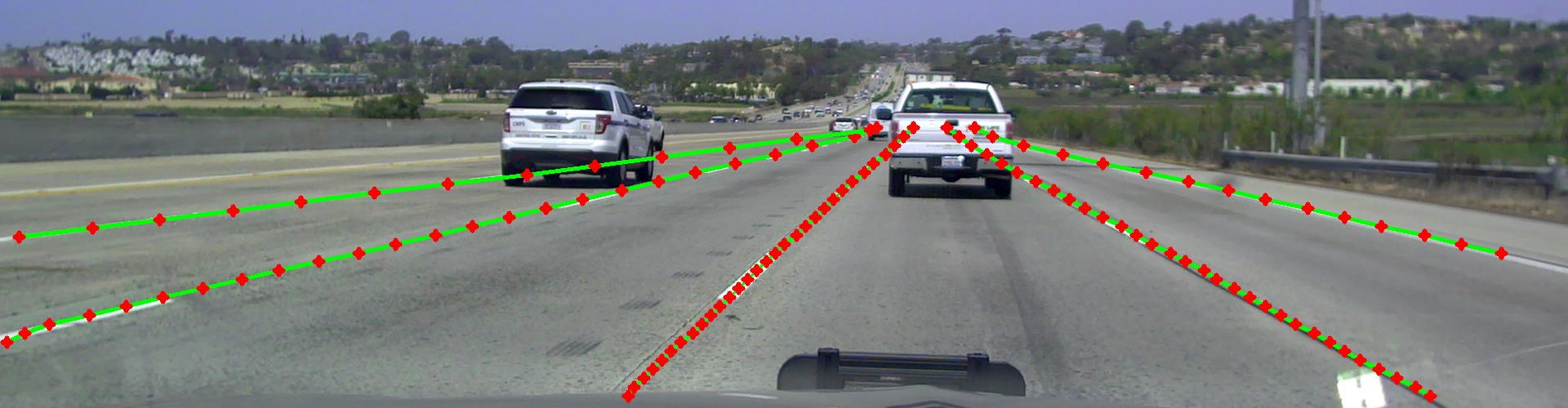}} \ &
	        \multicolumn{2}{@{}c@{}}{\includegraphics[width=0.48\linewidth,height=42px]{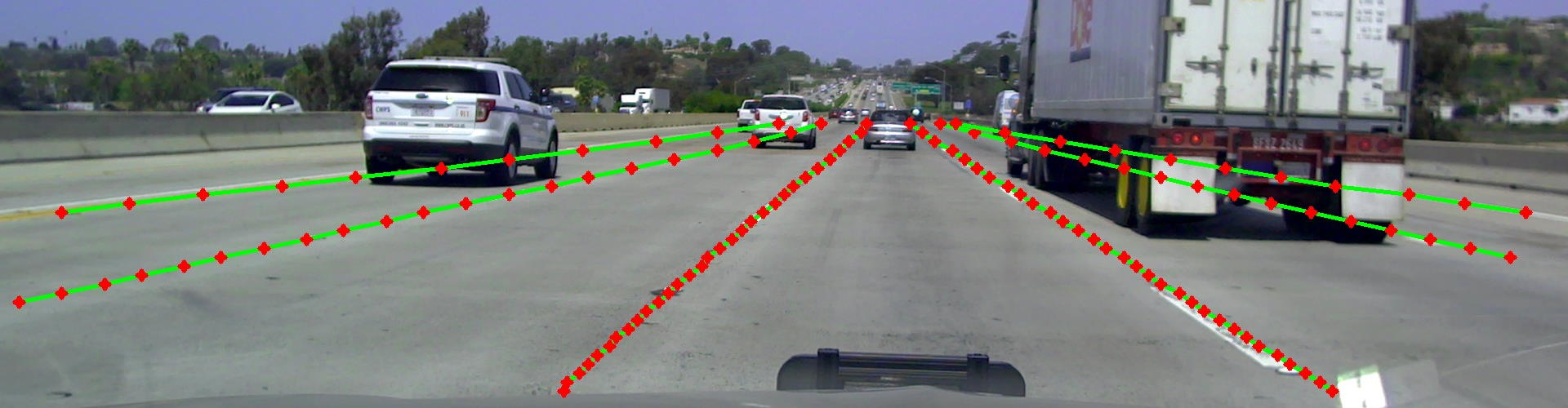}} \\
		    \includegraphics[width=.235\linewidth,height=72px]{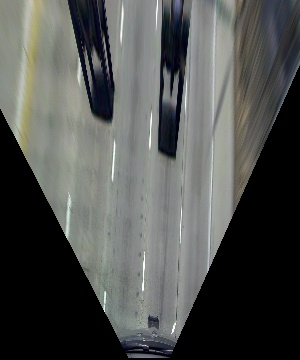} \ &
		    \includegraphics[width=.235\linewidth,height=72px]{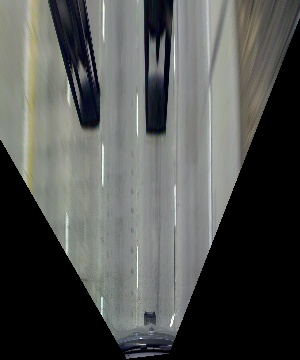} \ &
		    \includegraphics[width=.235\linewidth,height=72px]{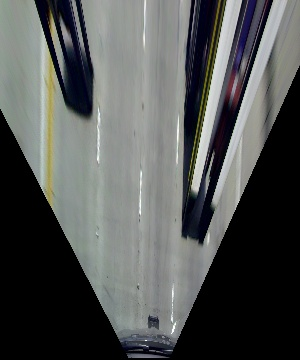} \ &
		    \includegraphics[width=.235\linewidth,height=72px]{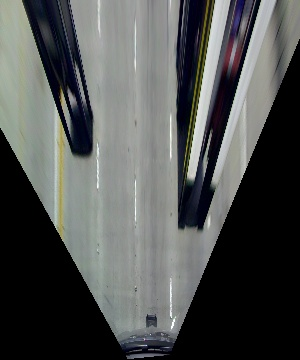}\\
		    \multicolumn{2}{@{}c@{}}{Frame \#201} &
		    \multicolumn{2}{@{}c@{}}{Frame \#772} \\
		    \multicolumn{4}{@{}c@{}}{\textit{Test2} dataset}
		\end{tabular} 
	\end{center} 
	\caption{Result of our online extrinsic camera calibration in the real-world scenes. In each triplet of images, the top, bottom-left, and bottom-right images show an input image, the BEV image based on given extrinsic camera parameters, and the BEV image based on those updated by the proposed method, respectively. In the input image, red vertices and green lines are from the lane boundary detection.} \vspace{-8pt}
	\label{fig:real_res} 
\end{figure}

In the real-world scenes, we compare the BEV images by IPM before and after applying the proposed method as shown in Fig.~\ref{fig:real_res}.
In each triplet of images in Fig.~\ref{fig:real_res}, the bottom-left and bottom-right images are the BEV images based on given extrinsic camera parameters and those updated by the proposed method, respectively.
The bottom-left images show that road surface fluctuates and lane widths are not equal, whereas the proposed method produces temporally consistent BEV images with the same lane widths and less fluctuation.

\begin{figure}[t]
    \begin{tabular}{@{}c@{}}
        \ \includegraphics[width=.95\linewidth,height=64px]{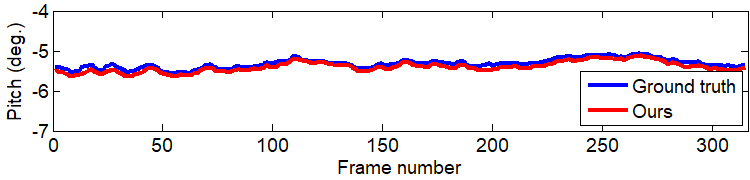} \\
        \ \includegraphics[width=.95\linewidth,height=64px]{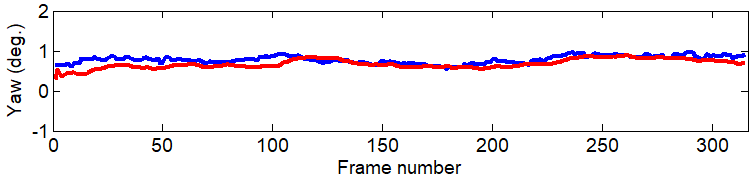} \\
        \ \ \includegraphics[width=.94\linewidth,height=64px]{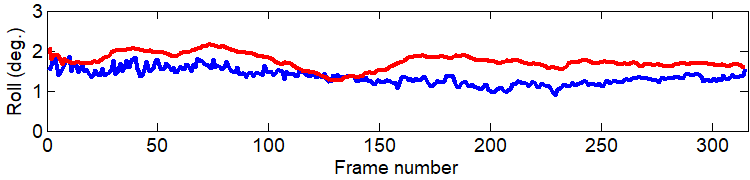} \\
        \includegraphics[width=.97\linewidth,height=64px]{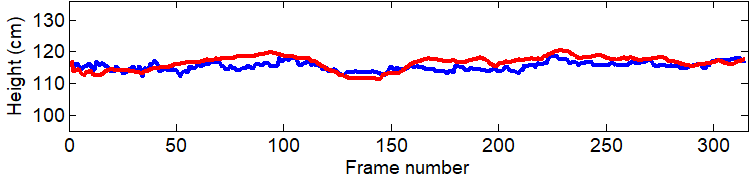}
    \end{tabular} 
    \caption{Extrinsic camera parameter estimates by our method on the \textit{Test1} dataset}
    \label{fig:real_graph}
\end{figure}

For quantitative comparison in the real-world scenes, we generated the pseudo ground truth of the extrinsic camera parameters in \textit{Test1} by optimizing (\ref{eq:meas_py}) and (\ref{eq:whenergy}) using manually annotated lane boundaries in each frame.
Fig.~\ref{fig:real_graph} shows the pseudo ground truth and the extrinsic camera parameter estimates by our method on the \textit{Test1} dataset. It is believed that the extrinsic camera parameter estimates are close to optimum although the pseudo ground truth may not be exactly the same as the real ground truth. 


\section{Conclusions}

In this paper, we proposed a method for online extrinsic camera calibration using lane boundary observations with the prior knowledge of lane width.
The proposed method estimated the geometric relationship between camera and road surface in two steps, \textit{i.e.}, we first estimated the VP-based pitch and yaw angle and then estimated roll angle and camera height using the pitch and yaw angles from the previous step based on the consistency of lane widths.
By virtue of EKF, our method finally produced temporally consistent BEV images with less fluctuation and equal lane widths.

\bibliographystyle{IEEEtran.bst}
\bibliography{IEEEabrv.bib}

\end{document}